\documentclass[conference,letterpaper]{IEEEtran}

\usepackage[utf8]{inputenc} 
\usepackage[T1]{fontenc}
\usepackage{url}
\usepackage{ifthen}
\usepackage{cite}
\usepackage[cmex10]{amsmath} 
\usepackage[capitalize,noabbrev]{cleveref}

\usepackage{algorithm}
\usepackage{algorithmic}
\usepackage{booktabs} 
\usepackage{multirow}
\usepackage{amsmath}
\usepackage{amssymb}
\usepackage{mathtools}
\usepackage{amsthm}
\usepackage{stfloats}
\newtheorem{corollary}{Corollary}

\newtheorem{theorem}{Theorem}
\newtheorem{lemma}{Lemma}

\newtheorem{assumption}{Assumption}

\usepackage{xcolor}
\definecolor{darkgreen}{RGB}{0,128,0}
\definecolor{darkred}{RGB}{200,0,0}
\usepackage{color, colortbl}
\usepackage{mwe}

\usepackage{algorithm,algorithmic}

\usepackage{multicol}
\usepackage{multirow}
\usepackage{arydshln}

\usepackage[capitalize,noabbrev]{cleveref}
\usepackage{newfloat}
\usepackage{listings}

\begin{document}
\title{Conditional Mutual Information Based Diffusion Posterior Sampling for Solving Inverse Problems} 

\author{%
  \IEEEauthorblockN{Shayan Mohajer Hamidi, En-Hui Yang}
  \IEEEauthorblockA{Department of Electrical and Computer Engineering \\
                    University of Waterloo,\
                    Waterloo, Canada,\
                    Email: \{smohajer,ehyang\}@uwaterloo.ca.}
}

\maketitle


\begin{abstract}
Inverse problems are prevalent across various disciplines in science and engineering. In the field of computer vision, tasks such as inpainting, deblurring, and super-resolution are commonly formulated as inverse problems. Recently, diffusion models (DMs) have emerged as a promising approach for addressing noisy linear inverse problems, offering effective solutions without requiring additional task-specific training. Specifically, with the prior provided by DMs, one can sample from the posterior by finding the likelihood. Since the likelihood is intractable, it is often approximated in the literature. However, this approximation compromises the quality of the generated images. To overcome this limitation and improve the effectiveness of DMs in solving inverse problems, we propose an information-theoretic approach. Specifically, we maximize the conditional mutual information $\mathrm{I}(\boldsymbol{x}_0; \boldsymbol{y} | \boldsymbol{x}_t)$, where $\boldsymbol{x}_0$ represents the reconstructed signal, $\boldsymbol{y}$ is the measurement, and $\boldsymbol{x}_t$ is the intermediate signal at stage $t$. This ensures that the intermediate signals $\boldsymbol{x}_t$ are generated in a way that the final reconstructed signal $\boldsymbol{x}_0$ retains as much information as possible about the measurement $\boldsymbol{y}$. We demonstrate that this method can be seamlessly integrated with recent approaches and, once incorporated, enhances their performance both qualitatively and quantitatively. 
\end{abstract}

\section{Introduction}
Diffusion models (DMs) are generative frameworks that progressively transform random noise into data by simulating a reverse diffusion process \cite{ddpm,song2019generative,songscore}. Starting with a noisy sample $\boldsymbol{x}_T$ from a Gaussian distribution, they iteratively refine it through intermediate states $\boldsymbol{x}_t$ ($t = T, \dots, 0$), using a learned score function that approximates the gradient of the log-density of the data distribution, $\nabla_{\boldsymbol{x}_t} \log p_t(\boldsymbol{x}_t)$.

Leveraging the remarkable capability of DMs to estimate target distributions, a promising application lies in utilizing them to address linear inverse problems, including tasks such as denoising, inpainting, deblurring, and super-resolution. These tasks involve recovering a signal $\boldsymbol{x}_0$ (e.g., a face image) from a measurement $\boldsymbol{y}$, where $\boldsymbol{y}$ is generated by a forward measurement operator $\boldsymbol{A}$ applied to $\boldsymbol{x}_0$, combined with measurement noise $\boldsymbol{n}$ \cite{song2020denoising,DPS, song2023pseudoinverse, dou2024diffusion,peng2024improving}. A straightforward approach to leverage DMs for solving inverse problems is to train a conditional DM to estimate the posterior $ p(\boldsymbol{x}_0 | \boldsymbol{y}) $. However, this method is computationally expensive, as it necessitates training different models for each measurement operator, significantly increasing the training overhead.

To address this challenge, recent approaches propose approximating the posterior by leveraging the estimated $\nabla_{\boldsymbol{x}_t} \log p_t(\boldsymbol{x}_t)$ obtained from DMs. This method eliminates the need for additional training, significantly reducing computational overhead. In this framework, $\nabla_{\boldsymbol{x}_t} \log p_t(\boldsymbol{x}_t)$ is combined with the likelihood gradient $\nabla_{\boldsymbol{x}_t} \log p_t(\boldsymbol{y} | \boldsymbol{x}_t)$ to compute the posterior gradient $\nabla_{\boldsymbol{x}_t} \log p_t(\boldsymbol{x}_t | \boldsymbol{y})$. This gradient enables sampling from the posterior distribution for solving inverse problems. However, due to the time-dependent nature of DMs, the likelihood term $p_t(\boldsymbol{y} | \boldsymbol{x}_t)$ is analytically intractable and must be approximated using alternative methods \cite{DPS,DSG,boys2023tweedie,hamididiffusion,song2023pseudoinverse}. These approximations often lead to samples drawn from the predicted distribution, $\boldsymbol{x}_t \sim p_t(\boldsymbol{x}_t | \boldsymbol{y})$, deviating from the true posterior distribution. Consequently, this misalignment can lead to either neglecting the measurement data, producing unrelated outputs, or generating images of low quality \cite{yangguidance}.

To address the aforementioned challenge and enhance the quality of reconstructed images using DMs, this paper adopts an information-theoretic approach to posterior sampling in DMs. Specifically, during each reverse step of the diffusion process, we aim to maximize the conditional mutual information (CMI), $\mathrm{I}(\boldsymbol{x}_0; \boldsymbol{y} | \boldsymbol{x}_t)$. By maximizing the CMI, we ensure that at each reverse step, $\boldsymbol{x}_t$ is generated in a way that the final reconstruction, $\boldsymbol{x}_0$, retains as much information as possible about the measurement $\boldsymbol{y}$. This approach mitigates the limitations of existing methods, where intermediate states $\boldsymbol{x}_t$ often deviate from the true posterior $p_t(\boldsymbol{x}_t | \boldsymbol{y})$, leading to suboptimal reconstructions.

To maximize the CMI, we take several steps. First, we derive a closed-form expression for $\mathrm{I} (\boldsymbol{x}_0; \boldsymbol{y} | \boldsymbol{x}_t)$ under commonly made assumptions in the DMs literature. Next, we compute a closed-form expression for the gradient of $\mathrm{I} (\boldsymbol{x}_0; \boldsymbol{y} | \boldsymbol{x}_t)$ with respect to (w.r.t.) $\boldsymbol{x}_t$, i.e., $\nabla_{\boldsymbol{x}_t} \mathrm{I} (\boldsymbol{x}_0; \boldsymbol{y} | \boldsymbol{x}_t)$. This gradient enables us to adjust $\boldsymbol{x}_t$ during the reverse diffusion process to progressively maximize $\mathrm{I} (\boldsymbol{x}_0; \boldsymbol{y} | \boldsymbol{x}_t)$.

We observe that the gradient $\nabla_{\boldsymbol{x}_t} \mathrm{I} (\boldsymbol{x}_0; \boldsymbol{y} | \boldsymbol{x}_t)$ involves the third-order derivative $\nabla^3_{\boldsymbol{x}_t} \log p_t(\boldsymbol{x}_t)$. Directly computing $\nabla^3_{\boldsymbol{x}_t} \log p_t(\boldsymbol{x}_t)$ from $\nabla_{\boldsymbol{x}_t} \log p_t(\boldsymbol{x}_t)$, which is estimated by DMs, introduces significant computational complexity. To overcome this challenge, we propose a method to approximate $\nabla_{\boldsymbol{x}_t} \mathrm{I} (\boldsymbol{x}_0; \boldsymbol{y} | \boldsymbol{x}_t)$ using Hutchinson’s trace estimation \cite{hutchinson1989stochastic}. This approach eliminates the need for explicitly computing the third-order derivative $\nabla^3_{\boldsymbol{x}_t} \log p_t(\boldsymbol{x}_t)$, thereby greatly reducing computational overhead.  Lastly, we demonstrate that our proposed method can be seamlessly integrated into existing approaches, such as DPS \cite{DPS}, $\Pi$GDM \cite{song2023pseudoinverse}, MCG \cite{MCG}, DSG \cite{DSG}, by adding just one line of code to their respective algorithms. Furthermore, through experiments conducted on two widely used datasets, FFHQ \cite{karras2019style} and ImageNet \cite{deng2009imagenet}, we show that incorporating our method enhances the performance of these approaches both qualitatively and quantitatively across a range of tasks, including inpainting, deblurring, and super-resolution.

\par{\textbf{Notation}:} We use the following notation throughout the paper. Scalars are represented by lowercase letters (e.g., $a$ or $A$), vectors by bold lowercase letters (e.g., $\boldsymbol{a}$), and matrices/tensors by bold uppercase letters (e.g., $\boldsymbol{A}$). The determinant and trace of a matrix $\boldsymbol{A}$ are represented by $\det(\boldsymbol{A})$ and $\mathrm{Tr}(\boldsymbol{A})$, respectively. The real axis is denoted by $\mathbb{R}$. The symbols $\bold{0}$ and $\bold{I}$ represent the zero and the identity matrix, respectively.

\section{Background and Preliminaries} \label{sec:pre}

\subsection{Diffusion Models}
Diffusion models define the generative process as the reverse of a noise addition process. Specifically, \cite{song2020score} introduced the Itô stochastic differential equation (SDE) to represent the noise addition process (i.e., the forward SDE) for the data $\boldsymbol{x}(t)$ over the time interval $t \in [0, T]$, where $\boldsymbol{x}(t) \in \mathbb{R}^d$ for all $t$.

In this paper, we adopt the variance-preserving stochastic differential equation (VP-SDE) \cite{song2020score}, which is equivalent to the denoising diffusion probabilistic model (DDPM) framework \cite{ddpm}. The corresponding equation is given as follows:
\begin{align}
\label{eq:forward-sde}
    d\boldsymbol{x} = -\frac{\beta(t)}{2} \boldsymbol{x} \, dt + \sqrt{\beta(t)}\, d\boldsymbol{w},
\end{align}
where $\beta(t)>0$ represents the noise schedule of the process \cite{ddpm}. The term $\boldsymbol{w}$ denotes the standard $d$-dimensional Wiener process. The data distribution is defined at $t=0$, i.e., $\boldsymbol{x}(0) \sim p_{\mathsf{data}}$, while a simple and tractable distribution, such as an isotropic Gaussian, is achieved at $t=T$, i.e., $\boldsymbol{x}(T) \sim \mathcal{N} (\bold{0}, \bold{I})$.

The objective is to recover the data-generating distribution from the tractable distribution. This is achieved by formulating the reverse SDE for \cref{eq:forward-sde} as follows \cite{anderson1982reverse}:
\begin{align}
\label{eq:reverse-sde}
    d\boldsymbol{x} = \left[-\frac{\beta(t)}{2} \boldsymbol{x} - \beta(t)\nabla_{\boldsymbol{x}_t} \log p_t({\boldsymbol{x}_t})\right] dt + \sqrt{\beta(t)} d\bar{\boldsymbol{w}},
\end{align}
where $dt$ represents time running backward, and $d\bar{\boldsymbol{w}}$ corresponds to the standard Wiener process running in reverse. The drift function now depends on the time-dependent score function $\nabla_{\boldsymbol{x}_t} \log p_t(\boldsymbol{x}_t)$, which is approximated by a neural network $\boldsymbol{s}_\theta$ trained using denoising score matching~\cite{vincent2011connection}. Specifically, this network takes $\boldsymbol{x}$ and $t$ as inputs to approximate the score function: $\boldsymbol{s}_\theta(\boldsymbol{x}, t) \approx \nabla_{\boldsymbol{x}_t} \log p_t(\boldsymbol{x}_t)$.

Discretizing \cref{eq:reverse-sde} and solving it produces samples from the data distribution $p(\boldsymbol{x}_0)$, which is the ultimate goal of generative modeling. In discrete settings with $N$ time steps, we define $\boldsymbol{x}_i \triangleq \boldsymbol{x}(iT/N)$ and $\beta_i \triangleq \beta(iT/N)$. Following \cite{ddpm}, we further introduce $\alpha_i \triangleq 1 - \beta_i$ and $\bar{\alpha}_i \triangleq \prod_{j=1}^i \alpha_j$.

\subsection{Diffusion Models for Solving Inverse Problems} \label{sec:setup}
We consider linear inverse problems for reconstructing an unknown signal $\boldsymbol{x}_0 \in \mathbb{R}^d$ from noisy measurements $\boldsymbol{y}\in \mathbb{R}^m$:
\begin{equation}
\label{eq:forward-model}
    \boldsymbol{y} = \bold{A}\boldsymbol{x}_0 + \boldsymbol{n},
\end{equation}
where $\bold{A}\in \mathbb{R}^{m\times d}$ is a known measurement operator and $\boldsymbol{n}\sim \mathcal{N}(\boldsymbol{0}, \boldsymbol{\Sigma}_{\boldsymbol{n}})$ represents additive Gaussian noise with a known covariance matrix $\boldsymbol{\Sigma}_{\boldsymbol{n}}$. This gives a likelihood function $p(\boldsymbol{y}|\boldsymbol{x}_0)=\mathcal{N}(\bold{A} \boldsymbol{x}_0, \boldsymbol{\Sigma}_{\boldsymbol{n}})$.

When $m < d$, which is common in real-world scenarios, the problem becomes ill-posed, requiring a \textit{prior} to obtain meaningful solutions. In the Bayesian framework, the prior $p(\boldsymbol{x}_0)$ is used to sample from the posterior $p(\boldsymbol{x}_0|\boldsymbol{y})$, defined by Bayes' rule: $p(\boldsymbol{x}_0|\boldsymbol{y}) = p(\boldsymbol{y}|\boldsymbol{x}_0)p(\boldsymbol{x}_0)/p(\boldsymbol{y})$. Using the diffusion model as the prior, \cref{eq:reverse-sde} can be adapted to create a reverse diffusion sampler for the posterior:
\begin{align} \label{eq:reverse-sde-posterior}
d \boldsymbol{x} = \Big[ & -\frac{\beta(t)}{2} \boldsymbol{x}  - \beta(t)(\nabla_{\boldsymbol{x}_t} \log p_t(\boldsymbol{x}_t) \nonumber \\
& + \nabla_{\boldsymbol{x}_t} \log p_t(\boldsymbol{y}|\boldsymbol{x}_t))\Big] dt + \sqrt{\beta(t)} d\bar{\boldsymbol{w}},
\end{align}
where we have used the fact that
\begin{align}
\label{eq:grad_log_bayes}
    \nabla_{\boldsymbol{x}_t} \log p_t(\boldsymbol{x}_t|\boldsymbol{y}) = \nabla_{\boldsymbol{x}_t} \log p_t(\boldsymbol{x}_t) + \nabla_{\boldsymbol{x}_t} \log p_t(\boldsymbol{y}|\boldsymbol{x}_t).
\end{align}
In \cref{eq:reverse-sde-posterior}, two terms must be computed: the score function \(\nabla_{\boldsymbol{x}_t} \log p_t(\boldsymbol{x}_t)\) and the likelihood \(\nabla_{\boldsymbol{x}_t} \log p_t(\boldsymbol{y}|\boldsymbol{x}_t)\). The former is directly obtained from the pre-trained score function \(\boldsymbol{s}_{\theta}\), while the latter is challenging to compute due to its time dependence, as the explicit relationship exists only between \(\boldsymbol{y}\) and \(\boldsymbol{x}_0\). Consequently, \(p_t(\boldsymbol{y}|\boldsymbol{x}_t)\) must be approximated, often resulting in samples that deviate from the measurements or exhibit poor quality.


\section{methodology} \label{sec:meth}
In this section, we present our methodology, which aims to enhance the quality of reconstructed images generated by DMs. Specifically, our approach focuses on maximizing the CMI value $\mathrm{I} (\boldsymbol{x}_0; \boldsymbol{y} | \boldsymbol{x}_t)$ during the reverse diffusion process. To achieve this, we proceed as follows: in \cref{sec:CMI}, we derive a closed-form expression for $\mathrm{I} (\boldsymbol{x}_0; \boldsymbol{y} | \boldsymbol{x}_t)$; in \cref{sec:derCMI}, we compute the gradient $\nabla_{\boldsymbol{x}_t} \mathrm{I} (\boldsymbol{x}_0; \boldsymbol{y} | \boldsymbol{x}_t)$ to facilitate its maximization; lastly, in \cref{sec:approx}, we propose an efficient method to approximate $\nabla_{\boldsymbol{x}_t} \mathrm{I} (\boldsymbol{x}_0; \boldsymbol{y} | \boldsymbol{x}_t)$, addressing computational challenges associated with its direct calculation.

\subsection{Calculating $\mathrm{I} (\boldsymbol{x}_0;\boldsymbol{y}|\boldsymbol{x}_t)$} \label{sec:CMI}
First, following the works of \cite{song2023pseudoinverse,peng2024improving,hamididiffusion,boys2023tweedie,rout2024beyond}, we introduce the following assumption:
\begin{assumption} \label{ass:Gauss}
The conditional distribution $p(\boldsymbol{x}_0|\boldsymbol{x}_t)$ follows a multivariate 
normal distribution, i.e., $p(\boldsymbol{x}_0|\boldsymbol{x}_t) \sim \mathcal{N} (\boldsymbol{\mu}_{\mathsf{post}} , \boldsymbol{\Sigma}_{\mathsf{post}})$.
\end{assumption}
Under this assumption, we derive the following theorem, which provides a closed-form expression for computing $\mathrm{I} (\boldsymbol{x}_0; \boldsymbol{y} | \boldsymbol{x}_t)$ in the context of DMs.
\begin{theorem}
Under Assumption 1, the conditional mutual information $\mathrm{I} (\boldsymbol{x}_0;\boldsymbol{y}|\boldsymbol{x}_t)$ is given by
\begin{align}
& \mathrm{I} (\boldsymbol{x}_0;\boldsymbol{y}|\boldsymbol{x}_t) =  \frac{1}{2}  \log \frac{\det \left( \boldsymbol{\Sigma}_{\mathsf{post}} \right)}{\det \left( \boldsymbol{\Sigma}_{\mathsf{post},\boldsymbol{y}} \right) },
\end{align}
where
\begin{subequations}
\begin{align}
&\boldsymbol{\Sigma}_{\mathsf{post}} = \frac{1-\bar\alpha_t}{\bar\alpha_t} \Big( \bold{I} + (1-\bar\alpha_t) \nabla^2_{\boldsymbol{x}_t}\log p_t(\boldsymbol{x}_t) \Big), \\
& \boldsymbol{\Sigma}_{\mathsf{post},\boldsymbol{y}} = \left[ \boldsymbol{\Sigma}_{\mathsf{post}}^{-1} + \bold{A}^{\mathsf{T}} \boldsymbol{\Sigma}_{\boldsymbol{n}}^{-1} \bold{A} \right]^{-1}.
\end{align}
\end{subequations}
\end{theorem}

\begin{proof}
We begin the proof by expressing $\mathrm{I} (\boldsymbol{x}_0;\boldsymbol{y}|\boldsymbol{x}_t)$ in terms of conditional differential entropy:
\begin{align} \label{eq:CMI}
\mathrm{I} (\boldsymbol{x}_0;\boldsymbol{y}|\boldsymbol{x}_t) = \mathrm{H} (\boldsymbol{x}_0|\boldsymbol{x}_t) - \mathrm{H} (\boldsymbol{x}_0| \boldsymbol{x}_t,\boldsymbol{y}).    
\end{align}
Next, we compute the two conditional differential entropy terms on the right-hand side of \cref{eq:CMI} separately.

\paragraph{Computing $\mathrm{H} (\boldsymbol{x}_0|\boldsymbol{x}_t)$}
As per the assumption in the theorem, $p(\boldsymbol{x}_0|\boldsymbol{x}_t) \sim \mathcal{N} (\boldsymbol{\mu}_{\mathsf{post}} , \boldsymbol{\Sigma}_{\mathsf{post}})$. In this case, the covariance $\boldsymbol{\Sigma}_{\mathsf{post}}$ is given by \cite{hamididiffusion,rout2024beyond}
\begin{align} \label{eq:covpost}
\boldsymbol{\Sigma}_{\mathsf{post}} = \frac{1-\bar\alpha_t}{\bar\alpha_t} \Big( \bold{I} + (1-\bar\alpha_t) \nabla^2_{\boldsymbol{x}_t}\log p_t(\boldsymbol{x}_t) \Big),    
\end{align}
where the Hessian $\nabla^2_{\boldsymbol{x}_t} \log p_t(\boldsymbol{x}_t)$ is obtained as $ \left[\nabla^2_{\boldsymbol{x}_t} \log p_t(\boldsymbol{x}_t)\right]_{ij} = \frac{\partial^2 \log p_t(\boldsymbol{x}_t)}{\partial x_{t,i} \partial x_{t,j}}$.
To calculate the entropy of $ p(\boldsymbol{x}_0|\boldsymbol{x}_t) $, we use the property that the entropy of a Gaussian random vector $ \boldsymbol{x} \sim \mathcal{N}(\boldsymbol{\mu}, \boldsymbol{\Sigma}) $ is given by $ \mathrm{H}(\boldsymbol{x}) = \frac{1}{2} \log \det(2\pi e \boldsymbol{\Sigma}) $. Using this, we obtain: 
\begin{align} \label{eq:ent1}
\mathrm{H} (\boldsymbol{x}_0|\boldsymbol{x}_t) = \frac{1}{2}  \log \det(2\pi e \boldsymbol{\Sigma}_{\mathsf{post}})  .
\end{align}

\paragraph{Computing $\mathrm{H} (\boldsymbol{x}_0| \boldsymbol{x}_t,\boldsymbol{y})$} To determine $ p(\boldsymbol{x}_0 | \boldsymbol{x}_t, \boldsymbol{y}) $, we note that the conditional prior $ p(\boldsymbol{x}_0 | \boldsymbol{x}_t) \sim \mathcal{N}(\boldsymbol{\mu}_{\mathsf{post}}, \boldsymbol{\Sigma}_{\mathsf{post}}) $ and the likelihood $ p(\boldsymbol{y} | \boldsymbol{x}_0) \sim \mathcal{N}(\bold{A} \boldsymbol{x}_0, \boldsymbol{\Sigma}_{\boldsymbol{n}}) $. According to the Gauss–Markov theorem \cite{gelman1995bayesian} (or the Kalman update rule \cite{kalman1960new}), the posterior $ p(\boldsymbol{x}_0 | \boldsymbol{x}_t, \boldsymbol{y}) $ is also Gaussian, with covariance matrix $ \boldsymbol{\Sigma}_{\mathsf{post},\boldsymbol{y}} $ given by:
\begin{align} \label{eq:ent2}
\boldsymbol{\Sigma}_{\mathsf{post},\boldsymbol{y}} = \left[ \boldsymbol{\Sigma}_{\mathsf{post}}^{-1} + \bold{A}^{\mathsf{T}} \boldsymbol{\Sigma}_{\boldsymbol{n}}^{-1} \bold{A} \right]^{-1}.   
\end{align}
Now, substituting \cref{eq:ent1} and \cref{eq:ent2} in \cref{eq:CMI} we obtain:
\begin{align}
\mathrm{I} (\boldsymbol{x}_0;\boldsymbol{y}|\boldsymbol{x}_t) & =  \frac{1}{2}  \log \det(2\pi e \boldsymbol{\Sigma}_{\mathsf{post}})    - \frac{1}{2}  \log \det(2\pi e \boldsymbol{\Sigma}_{\mathsf{post},\boldsymbol{y}}) \nonumber \\
& =  \frac{1}{2}  \log \frac{\det(\boldsymbol{\Sigma}_{\mathsf{post}})}{\det(\boldsymbol{\Sigma}_{\mathsf{post},\boldsymbol{y}})}, 
\end{align}
which concludes the theorem. 
\end{proof}

\subsection{Computing $\nabla_{\boldsymbol{x}_t}  \mathrm{I} (\boldsymbol{x}_0;\boldsymbol{y}|\boldsymbol{x}_t)$} \label{sec:derCMI}

Now, our objective is to take a gradient step to maximize the CMI value $\mathrm{I} (\boldsymbol{x}_0;\boldsymbol{y}|\boldsymbol{x}_t)$ during each reverse step of the diffusion process. To achieve this, the following theorem provides the gradient of $\mathrm{I}(\boldsymbol{x}_0; \boldsymbol{y} | \boldsymbol{x}_t)$ w.r.t. $\boldsymbol{x}_t$. Before we introduce the theorem, we explain two notations we use hereafter:

Assume that $\bold{E} \in \mathbb{R}^{d \times d}$ is a two-dimensional matrix and $\bold{F} \in \mathbb{R}^{d \times d \times d}$ is a three-dimensional tensor. The operations  $ \bold{E} \odot_1 \bold{F}$ and $\bold{F} \odot_2 \bold{E} $ are defined as follows:
\begin{itemize}
\item The operation $ \bold{E} \odot_1 \bold{F}$ represents the contraction of a 2D matrix $\bold{E}$ with the 3D tensor $\bold{F}$ along the third dimension of the tensor $\bold{F}$, i.e., $\left[  \bold{E}  \odot_1 \bold{F}  \right]_{:,:,k}  = \bold{E} \left[\bold{F}\right]_{:,:,k}$, which results in a $d \times d \times d$ tensor. 

\item The operation $\bold{F} \odot_2 \bold{E} $ represents a contraction where the 3D tensor $\bold{F}$ is multiplied along its third dimension by the 2D matrix $\bold{E}$, i.e., $\left[  \bold{F}  \odot_2 \bold{E}  \right]_{:,:,k}  =  \left[\bold{F}\right]_{:,:,k} \bold{E}$, which results in a $d \times d \times d$ tensor. 
\end{itemize}
Now, with these two operations at hand, we proceed to introduce \cref{th:der}.

\begin{theorem} \label{th:der}
The gradient of $\mathrm{I} (\boldsymbol{x}_0;\boldsymbol{y}|\boldsymbol{x}_t)$ w.r.t. $\boldsymbol{x}_t$ is given by
\begin{align}
\nabla_{\boldsymbol{x}_t}  \mathrm{I} (\boldsymbol{x}_0;\boldsymbol{y}|\boldsymbol{x}_t) & = \frac{1}{2} \Big[ \mathrm{Tr} \left( \boldsymbol{\Sigma}_{\mathsf{post}}^{-1}  \odot_1 \nabla_{\boldsymbol{x}_t} \boldsymbol{\Sigma}_{\mathsf{post}} \right) \nonumber \\ \label{eq:der-I}
& -  \mathrm{Tr} \left( \boldsymbol{\Sigma}_{\mathsf{post},\boldsymbol{y}}^{-1}  \odot_1 \nabla_{\boldsymbol{x}_t} \boldsymbol{\Sigma}_{\mathsf{post},\boldsymbol{y}} \right) \Big],
\end{align}
where the trace $\mathrm{Tr}(\cdot)$ is applied across the first two dimensions (matrix dimensions) for each slice along the third dimension, resulting in a $d$-dimensional vector for $\nabla_{\boldsymbol{x}_t}  \mathrm{I} (\boldsymbol{x}_0;\boldsymbol{y}|\boldsymbol{x}_t)$.

Also, the gradients $\nabla_{\boldsymbol{x}_t} \boldsymbol{\Sigma}_{\mathsf{post}}$ and $\nabla_{\boldsymbol{x}_t} \boldsymbol{\Sigma}_{\mathsf{post},\boldsymbol{y}}$ are computed as follows
\begin{subequations}
\begin{align} \label{eq:grad1}
\nabla_{\boldsymbol{x}_t} \boldsymbol{\Sigma}_{\mathsf{post}}  &=  \frac{(1-\bar\alpha_t)^2}{\bar\alpha_t} \nabla^3_{\boldsymbol{x}_t}\log p_t(\boldsymbol{x}_t),  \\ \nonumber
\nabla_{\boldsymbol{x}_t} \boldsymbol{\Sigma}_{\mathsf{post},\boldsymbol{y}} &= \boldsymbol{\Sigma}_{\mathsf{post},\boldsymbol{y}}  \odot_1 \Big[ \boldsymbol{\Sigma}_{\mathsf{post}}^{-1}  \odot_1 \nabla_{\boldsymbol{x}_t} \boldsymbol{\Sigma}_{\mathsf{post}}  \\ \label{eq:grad2}
& \qquad \qquad \odot_2 \boldsymbol{\Sigma}_{\mathsf{post}}^{-1}  \Big] \odot_2 \boldsymbol{\Sigma}_{\mathsf{post},\boldsymbol{y}},
\end{align}
\end{subequations}
where the entries of 3D tensor $\nabla^3_{\boldsymbol{x}_t}\log p_t(\boldsymbol{x}_t)$  is calculated as $\left[\nabla^3_{\boldsymbol{x}_t} \log p_t(\boldsymbol{x}_t)\right]_{ijk} = \frac{\partial}{\partial x_{t,k}} \left[\nabla^2_{\boldsymbol{x}_t} \log p_t(\boldsymbol{x}_t)\right]_{ij}$.
\end{theorem}

\begin{proof}
For the proof, we use the following Lemma. 
\begin{lemma} [Derivative of the log-determinant w.r.t. a parameter \cite{petersen2008matrix,magnus2019matrix}] \label{lemma:det} 
Let $ \bold{M}_{\theta} \in \mathbb{R}^{d \times d} $ be a square, symmetric, and positive-definite matrix that depends smoothly on a parameter $ \boldsymbol{\theta} \in \mathbb{R}^{d}$. The gradient of $ \log \det(\bold{M}_{\boldsymbol{\theta}}) $ w.r.t. $ \boldsymbol{\theta} $ is:
\[
\nabla_{\boldsymbol{\theta}} \log \det \left( \bold{M}_{\boldsymbol{\theta}} \right) = \mathrm{Tr} \left( \bold{M}_{\boldsymbol{\theta}}^{-1} \odot_1 \nabla_{\boldsymbol{\theta}} \bold{M}_{\boldsymbol{\theta}} \right).
\]
\end{lemma}
Using \cref{lemma:det}, it is easy to see that \cref{eq:der-I} holds. 

In addition, from \cref{eq:covpost}, and noting that the term $\frac{(1-\bar\alpha_t)^2}{\bar\alpha_t} $ is constant w.r.t. $\boldsymbol{x}_t$, it is straight-forward to check that the gradients $\nabla_{\boldsymbol{x}_t} \boldsymbol{\Sigma}_{\mathsf{post}}$ follows \cref{eq:grad1}.

On the other hand, to find $\nabla_{\boldsymbol{x}_t} \boldsymbol{\Sigma}_{\mathsf{post},\boldsymbol{y}}$, we first define $\bold{R} = \boldsymbol{\Sigma}_{\mathsf{post}}^{-1} + \bold{A}^{\mathsf{T}} \boldsymbol{\Sigma}_{\boldsymbol{n}}^{-1} \bold{A}$. Then, 
\begin{align} \label{eq:gradR}
\nabla_{\boldsymbol{x}_t} \bold{R} &=  \nabla_{\boldsymbol{x}_t} \left[ \boldsymbol{\Sigma}_{\mathsf{post}}^{-1} + \bold{A}^{\mathsf{T}} \boldsymbol{\Sigma}_{\boldsymbol{n}}^{-1} \bold{A} \right]  = \nabla_{\boldsymbol{x}_t} \boldsymbol{\Sigma}_{\mathsf{post}}^{-1},
\end{align}
where \cref{eq:gradR} is obtained since $\bold{A} \boldsymbol{\Sigma}_{\boldsymbol{n}}^{-1} \bold{A}$ is independent of $\boldsymbol{x}_t$ as both $\bold{A}$ and $\boldsymbol{\Sigma}_{\boldsymbol{n}}$ are independent of $\boldsymbol{x}_t$. Now, to find $\nabla_{\boldsymbol{x}_t} \boldsymbol{\Sigma}_{\mathsf{post},\boldsymbol{y}}$, we need to find $\nabla_{\boldsymbol{x}_t} \bold{R}^{-1}$. Toward this end, we use the following Lemma:
\begin{lemma} [Gradient of the Inverse of a Matrix \cite{petersen2008matrix}]
\label{lemma:matrix_grad}  
Let $ \bold{M}_{\boldsymbol{\theta}} \in \mathbb{R}^{d \times d} $ be an invertible matrix that depends smoothly on a parameter vector $ \boldsymbol{\theta} \in \mathbb{R}^p $. Then, the gradient of the inverse matrix $ \bold{M}_{\boldsymbol{\theta}}^{-1} $ w.r.t. $ \boldsymbol{\theta} $ is given by:
\[
\nabla_{\boldsymbol{\theta}} \bold{M}_{\boldsymbol{\theta}}^{-1} = - \bold{M}_{\boldsymbol{\theta}}^{-1} \odot_1 \nabla_{\boldsymbol{\theta}} \bold{M}_{\boldsymbol{\theta}} \odot_2 \bold{M}_{\boldsymbol{\theta}}^{-1},
\]
where $ \nabla_{\boldsymbol{\theta}} \bold{M}_{\boldsymbol{\theta}} \in \mathbb{R}^{d \times d \times p} $ is the gradient of $ \bold{M}_{\boldsymbol{\theta}} $ w.r.t. $ \boldsymbol{\theta} $, and $ \nabla_{\boldsymbol{\theta}} \bold{M}_{\boldsymbol{\theta}}^{-1} \in \mathbb{R}^{d \times d \times p} $ is the gradient of the inverse matrix.
\end{lemma}
Based on \cref{lemma:matrix_grad}, we can now find $\nabla_{\boldsymbol{x}_t}  \bold{R}^{-1}$  as follows
\begin{align}
& \nabla_{\boldsymbol{x}_t}  \bold{R}^{-1} = -   \bold{R}^{-1} \odot_1 \nabla_{\boldsymbol{x}_t} \bold{R} \odot_2 \bold{R}^{-1} \\
& = - \boldsymbol{\Sigma}_{\mathsf{post},\boldsymbol{y}}  \odot_1 \nabla_{\boldsymbol{x}_t} \boldsymbol{\Sigma}_{\mathsf{post}}^{-1} \odot_2 \boldsymbol{\Sigma}_{\mathsf{post},\boldsymbol{y}} \\ \label{eq:fromLemma2}
& = - \boldsymbol{\Sigma}_{\mathsf{post},\boldsymbol{y}}  \odot_1 \left[ -\boldsymbol{\Sigma}_{\mathsf{post}}^{-1} \odot_1 \nabla_{\boldsymbol{x}_t} \boldsymbol{\Sigma}_{\mathsf{post}} \odot_2 \boldsymbol{\Sigma}_{\mathsf{post}}^{-1} \right] \odot_2 \boldsymbol{\Sigma}_{\mathsf{post},\boldsymbol{y}}  \\ \label{eq:minuscancels}
& = \boldsymbol{\Sigma}_{\mathsf{post},\boldsymbol{y}}  \odot_1 \left[ \boldsymbol{\Sigma}_{\mathsf{post}}^{-1}  \odot_1 \nabla_{\boldsymbol{x}_t} \boldsymbol{\Sigma}_{\mathsf{post}}  \odot_2 \boldsymbol{\Sigma}_{\mathsf{post}}^{-1}  \right] \odot_2 \boldsymbol{\Sigma}_{\mathsf{post},\boldsymbol{y}}, 
\end{align}
where \cref{eq:fromLemma2} is obtained using \cref{lemma:matrix_grad} to calculate $\nabla_{\boldsymbol{x}_t} \boldsymbol{\Sigma}_{\mathsf{post}}^{-1}$, and \cref{eq:minuscancels} is obtained as double negative cancels. This concludes the Theorem.
\end{proof}

As described in \cref{eq:grad1}, the gradient $\nabla_{\boldsymbol{x}_t} \mathrm{I}(\boldsymbol{x}_0; \boldsymbol{y} | \boldsymbol{x}_t)$ involves the two terms $\mathrm{Tr} \left( \boldsymbol{\Sigma}_{\mathsf{post}}^{-1} \odot_1 \nabla_{\boldsymbol{x}_t} \boldsymbol{\Sigma}_{\mathsf{post}} \right)$ and $\mathrm{Tr} \left( \boldsymbol{\Sigma}_{\mathsf{post},\boldsymbol{y}}^{-1} \odot_1 \nabla_{\boldsymbol{x}_t} \boldsymbol{\Sigma}_{\mathsf{post},\boldsymbol{y}} \right)$, both of which depend on $\nabla^3_{\boldsymbol{x}_t} \log p_t(\boldsymbol{x}_t)$. In many diffusion models, such as denoising DMs, the score function $\nabla_{\boldsymbol{x}_t} \log p_t(\boldsymbol{x}_t)$ is readily available. As such, a naive approach to compute $\nabla^3_{\boldsymbol{x}_t} \log p_t(\boldsymbol{x}_t)$ would involve differentiating the score function $\nabla_{\boldsymbol{x}_t} \log p_t(\boldsymbol{x}_t)$ twice w.r.t. $\boldsymbol{x}_t$. However, for high-dimensional data such as images, where the dimensionality $d$ is large—as is often the case in the experimental section—this direct computation incurs a substantial computational burden due to the complexity of handling rank-3 tensors. 

To address this challenge, the next subsection introduces an efficient method to approximate the terms $\mathrm{Tr} \left( \boldsymbol{\Sigma}_{\mathsf{post}}^{-1} \odot_1 \nabla_{\boldsymbol{x}_t} \boldsymbol{\Sigma}_{\mathsf{post}} \right)$ and $\mathrm{Tr} \left( \boldsymbol{\Sigma}_{\mathsf{post},\boldsymbol{y}}^{-1} \odot_1 \nabla_{\boldsymbol{x}_t} \boldsymbol{\Sigma}_{\mathsf{post},\boldsymbol{y}} \right)$, thereby avoiding the direct computation of $\nabla^3_{\boldsymbol{x}_t} \log p_t(\boldsymbol{x}_t)$.

\subsection{Approximating $\mathrm{Tr} \left( \boldsymbol{\Sigma}_{\mathsf{post}}^{-1} \odot_1 \nabla_{\boldsymbol{x}_t} \boldsymbol{\Sigma}_{\mathsf{post}} \right)$ and $\mathrm{Tr} \left( \boldsymbol{\Sigma}_{\mathsf{post},\boldsymbol{y}}^{-1} \odot_1 \nabla_{\boldsymbol{x}_t} \boldsymbol{\Sigma}_{\mathsf{post},\boldsymbol{y}} \right)$} \label{sec:approx}

In this subsection, we detail the method to approximate $\mathrm{Tr} \left( \boldsymbol{\Sigma}_{\mathsf{post}}^{-1} \odot_1 \nabla_{\boldsymbol{x}_t} \boldsymbol{\Sigma}_{\mathsf{post}} \right)$. The procedure for approximating $\mathrm{Tr} \left( \boldsymbol{\Sigma}_{\mathsf{post},\boldsymbol{y}}^{-1} \odot_1 \nabla_{\boldsymbol{x}_t} \boldsymbol{\Sigma}_{\mathsf{post},\boldsymbol{y}} \right)$ follows a similar approach and is omitted here for brevity.

We first begin by simplifying $\mathrm{Tr} \left( \boldsymbol{\Sigma}_{\mathsf{post}}^{-1} \odot_1 \nabla_{\boldsymbol{x}_t} \boldsymbol{\Sigma}_{\mathsf{post}} \right)$. Using \cref{eq:grad1}, we have 
\begin{align}
\mathrm{Tr} & \left( \boldsymbol{\Sigma}_{\mathsf{post}}^{-1} \odot_1 \nabla_{\boldsymbol{x}_t} \boldsymbol{\Sigma}_{\mathsf{post}} \right) \nonumber \\
&= \frac{(1-\bar\alpha_t)^2}{\bar\alpha_t}  \mathrm{Tr} \left( \boldsymbol{\Sigma}_{\mathsf{post}}^{-1}  \odot_1 \nabla^3_{\boldsymbol{x}_t} \log p_t(\boldsymbol{x}_t) \right).   
\end{align}
Next, we use Hutchinson’s expansion to approximate the term $\mathrm{Tr} \left( \boldsymbol{\Sigma}_{\mathsf{post}}^{-1} \odot_1 \nabla^3_{\boldsymbol{x}_t} \log p_t(\boldsymbol{x}_t) \right)$ without the need for calculating the third-order derivative $\nabla^3_{\boldsymbol{x}_t} \log p_t(\boldsymbol{x}_t)$. We first, explain the Hutchinson’s expansion in the following Lemma.

\begin{lemma} [Hutchinson’s trace estimation \cite{hutchinson1989stochastic}] 
Let $ A \in \mathbb{R}^{n \times n} $ be a square matrix, and let $ \boldsymbol{v} \in \mathbb{R}^n $ be a random vector whose entries $ v_i $ are independent and satisfy the following properties:
\[
\mathbb{E}[v_i] = 0, \quad \mathbb{E}[v_i^2] = 1, \quad \text{and} \quad \mathbb{E}[v_i v_j] = 0 \quad \text{for } i \neq j.
\]
Then the trace of $ A $ can be expressed as:
\[
\mathrm{Tr}(A) = \mathbb{E}[\boldsymbol{v}^\top A \boldsymbol{v}],
\]
where the expectation is taken w.r.t. the distribution of $ \boldsymbol{v} $.  
\end{lemma}

\begin{corollary} \label{cor:tr}
In practice, the trace can be approximated using $ r $ independent samples of $ \boldsymbol{v} $ as: $\hat{\mathrm{Tr}}(A) = \frac{1}{r} \sum_{i=1}^r \boldsymbol{v}_i^\top A \boldsymbol{v}_i $,
where $ \boldsymbol{v}_i $ are independent random vectors. The estimator $ \hat{\mathrm{Tr}}(A) $ is unbiased, i.e., $\mathbb{E}[\hat{\mathrm{Tr}}(A)] = \mathrm{Tr}(A)$.    
\end{corollary}

Based on \cref{cor:tr}, the $k$-th entry of $\mathrm{Tr} \left( \boldsymbol{\Sigma}_{\mathsf{post}}^{-1} \odot_1 \nabla^3_{\boldsymbol{x}_t} \log p_t(\boldsymbol{x}_t) \right)$ can be calculated as follows
\begin{align} 
\big[ \mathrm{Tr} & \left( \boldsymbol{\Sigma}_{\mathsf{post}}^{-1}  \nabla^3_{\boldsymbol{x}_t} \log p_t(\boldsymbol{x}_t) \right) \big]_k \nonumber \\ \label{eq:approx}
&\approx   \frac{1}{r} \sum_{i=1}^r \boldsymbol{v}_i^\top \left( \boldsymbol{\Sigma}_{\mathsf{post}}^{-1}  \left[\nabla^3_{\boldsymbol{x}_t} \log p_t(\boldsymbol{x}_t) \right]_{k,:,:} \right) \boldsymbol{v}_i  \\ \label{eq:approx2}
&\approx   \frac{1}{r} \sum_{i=1}^r \boldsymbol{v}_i^\top  \boldsymbol{\Sigma}_{\mathsf{post}}^{-1}  \left( \left[\nabla^3_{\boldsymbol{x}_t} \log p_t(\boldsymbol{x}_t)  \right]_{k,:,:} \boldsymbol{v}_i \right),
\end{align}
where \cref{eq:approx2} is obtained due to associativity of matrix multiplication. 

Now, note that we can calculate the term in \cref{eq:approx2} without the need to explicitly calculate the 3D tensor $\nabla^3_{\boldsymbol{x}_t} \log p_t(\boldsymbol{x}_t)$; but instead, we only need to calculate $\nabla^3_{\boldsymbol{x}_t} \log p_t(\boldsymbol{x}_t) \boldsymbol{v}_i$ which is a 1D vector. In the sequel, we discuss how we can find $\nabla^3_{\boldsymbol{x}_t} \log p_t(\boldsymbol{x}_t) \boldsymbol{v}_i$.

The score function $\nabla_{\boldsymbol{x}_t} \log p_t(\boldsymbol{x}_t)$ is readily available from DMs. Using autodiff library (PyTorch, JAX, TensorFlow, etc.), we find the Hessian vector product $\nabla^2_{\boldsymbol{x}_t} \log p_t(\boldsymbol{x}_t)  \boldsymbol{v}_i$, i.e., the directional derivative of $\nabla_{\boldsymbol{x}_t} \log p_t(\boldsymbol{x}_t)$ in direction of $\boldsymbol{v}_i$. Note that $\nabla^2_{\boldsymbol{x}_t} \log p_t(\boldsymbol{x}_t)  \boldsymbol{v}_i$ is a $d$-dimensional vector, as opposed to the Hessian which is $d \times d$ matrix. 
Differentiating the Hessian-vector product  in direction $\boldsymbol{v}_i$ once more gives $\nabla^3_{\boldsymbol{x}_t} \log p_t(\boldsymbol{x}_t) \boldsymbol{v}_i$ which is a $d$-dimensional vector. This is far less memory-intensive than explicitly building a $\mathcal{O} (d^2)$ Hessian or $\mathcal{O} (d^3)$ third-order derivative.

Once we obtain $\nabla^3_{\boldsymbol{x}_t} \log p_t(\boldsymbol{x}_t)\boldsymbol{v}_i$, we plug it in \cref{eq:approx2} to obtain an approximation for $\mathrm{Tr} \left( \boldsymbol{\Sigma}_{\mathsf{post}}^{-1}  \nabla^3_{\boldsymbol{x}_t} \log p_t(\boldsymbol{x}_t) \right)$.

It is worth noting that our CMI maximization can be a plugin module to existing benchmark methods, such as DPS \cite{DPS}, $\Pi$GDM \cite{song2023pseudoinverse}, MCG \cite{MCG}, DSG \cite{DSG}. Notably, our method only requires modifying a few lines of code. Using DPS as an example, the algorithm for CMI+DPS is summarized in \cref{alg}.

\begin{algorithm}[H]
   \caption{CMI+DPS}
   \label{alg}
\begin{algorithmic}[1]
   \STATE   {\bfseries Input:} The number of iterations $N$, $\boldsymbol{y}$, noise levels $\{ \Tilde{\sigma}_t \}$, step-sizes $\{\eta_t\}$ and $\{\zeta_t\}$.  
   \STATE $\boldsymbol{x}_N \sim \mathcal{N} (\bold{0}, \bold{I})$
   \FOR{$t=N-1, N-2,\dots,0$}
   \STATE  $\hat{\boldsymbol{s}} \leftarrow \boldsymbol{s}_{\theta} (\boldsymbol{x}_t,t)$ 
   \STATE $ \Tilde{\boldsymbol{x}}_0 \leftarrow \frac{1}{\sqrt{\bar\alpha_t}}\left(\boldsymbol{x}_t+(1-\bar\alpha_t)\hat{\boldsymbol{s}} \right)$
   \STATE $\boldsymbol{z} \sim \mathcal{N} (\bold{0}, \bold{I})$.
   \STATE  $\boldsymbol{x}_{t-1}^\prime \leftarrow \frac{\sqrt{\alpha_t}(1-\bar{\alpha}_{t-1})}{1 - \bar{\alpha}_t}\boldsymbol{x}_t + \frac{\sqrt{\bar{\alpha}_{t-1}}\beta_t}{1 - \bar{\alpha}_t}\Tilde{\boldsymbol{x}}_0 +  {\tilde{\sigma}_t \boldsymbol{z}}$.
   \STATE  \color{blue}  
   $\boldsymbol{x}_{t-1}^\prime \leftarrow \boldsymbol{x}_{t-1}^\prime + \eta_t \nabla_{\boldsymbol{x}_t}  \mathrm{I} (\boldsymbol{x}_0;\boldsymbol{y}|\boldsymbol{x}_t)$ \hfill \textcolor{gray}{\# CMI step}
   \color{black}
   \STATE $\boldsymbol{x}_{t-1} \leftarrow \boldsymbol{x}_{t-1}^\prime - \zeta_t \nabla_{\boldsymbol{x}_{t}} \| \boldsymbol{y} -\bold{A}( \Tilde{\boldsymbol{x}}_0) \|$ \hfill \textcolor{gray}{\# DPS step}
   \ENDFOR
   \STATE  {\bfseries Output:} $\boldsymbol{x}_0$
\end{algorithmic}
\end{algorithm}

\begin{table*}[!t]
\renewcommand{\baselinestretch}{1.0}
\renewcommand{\arraystretch}{1.0}
\setlength{\tabcolsep}{2.2pt}
\caption{\normalfont Quantitative results on the $1k$ validation sets of FFHQ ($256\times256$) and ImageNet ($256\times256$) datasets. Results are highlighted in \textcolor{darkgreen}{green} to indicate improved performance when CMI is integrated with the baseline method, and in \textcolor{darkred}{red} to indicate a decrease in performance.}\label{tab:quant}
\vskip -0.1in
\centering
\resizebox{1\textwidth}{!}{\begin{tabular}{@{}c|cccccccccccccccc@{}}
\toprule
\multirow{2}{*}{\textbf{Dataset}} & \multirow{2}{*}{\textbf{Method}}      & \multicolumn{3}{c}{\textbf{Inpaint~(Random)}} & \multicolumn{3}{c}{\textbf{Inpaint~(Box)}} & \multicolumn{3}{c}{\textbf{Deblur~(Gaussian)}} & \multicolumn{3}{c}{\textbf{Deblur~(Motion)}} & \multicolumn{3}{c}{\textbf{SR~($4\times$)}} \\
\cmidrule(lr){3-5}  \cmidrule(lr){6-8}  \cmidrule(lr){9-11}  \cmidrule(lr){12-14} \cmidrule(lr){15-17}
&      & FID~$\downarrow$ & LPIPS~$\downarrow$ & SSIM~$\uparrow$ & FID~$\downarrow$ & LPIPS~$\downarrow$ & SSIM~$\uparrow$ & FID~$\downarrow$ & LPIPS~$\downarrow$ & SSIM~$\uparrow$ & FID~$\downarrow$ & LPIPS~$\downarrow$ & SSIM~$\uparrow$ & FID~$\downarrow$ & LPIPS~$\downarrow$ & SSIM~$\uparrow$  \\
\midrule
\multirow{7}{*}{FFHQ}  & DPS \cite{DPS} & 21.19 & 0.212 & 0.851 & 33.12 & 0.168& 0.873 & 44.05 & 0.257 &0.811 & 39.92 & 0.242 & 0.859 & 39.35 & 0.214 & 0.852 \\ 
& CMI+DPS (ours) &  \textcolor{darkgreen}{20.46} & \textcolor{darkgreen}{0.207} & \textcolor{darkgreen}{0.866} & \textcolor{darkgreen}{31.52} & \textcolor{darkgreen}{0.154} & \textcolor{darkgreen}{0.875} & \textcolor{darkgreen}{42.81} & \textcolor{darkgreen}{0.241} & \textcolor{darkgreen}{0.827} & \textcolor{darkgreen}{36.51} & \textcolor{darkgreen}{0.232} & \textcolor{darkgreen}{0.877} & \textcolor{darkgreen}{35.24} & \textcolor{darkgreen}{0.209} & \textcolor{darkgreen}{0.854}  \\ \cdashline{2-17}
&  $\Pi$GDM \cite{song2023pseudoinverse} & 21.27 & 0.221 & 0.840 & 34.79 & 0.179& 0.860 & 40.21 & 0.242&0.825 & 33.24 & 0.221 & 0.887 & 34.98 & 0.202 & 0.854  \\ 
& CMI+$\Pi$GDM (ours)& \textcolor{darkgreen}{20.55} & \textcolor{darkgreen}{0.215} & \textcolor{darkgreen}{0.869} & \textcolor{darkgreen}{32.59} & \textcolor{darkgreen}{0.148} & \textcolor{darkgreen}{0.871} & \textcolor{darkgreen}{37.73} & \textcolor{darkgreen}{0.240} & \textcolor{darkgreen}{0.832} & \textcolor{darkgreen}{31.20} & \textcolor{darkgreen}{0.217} & \textcolor{darkgreen}{0.897} & \textcolor{darkgreen}{32.74} & \textcolor{darkgreen}{0.200} & \textcolor{darkgreen}{0.857}  \\ \cdashline{2-17}
& MCG \cite{MCG} & 29.26 & 0.286 &  0.751 & 40.11 &0.309& 0.703&  101.2 &0.340&0.051& $-$ & $-$ & $-$ &87.64 &0.520  &0.559\\ 
& CMI+MCG (ours)& \textcolor{darkgreen}{26.35} & \textcolor{darkgreen}{0.267} &  \textcolor{darkgreen}{0.782} & \textcolor{darkgreen}{36.54} &\textcolor{darkgreen}{0.251}& \textcolor{darkgreen}{0.727}&  \textcolor{darkgreen}{85.18} &\textcolor{darkgreen}{0.325}&\textcolor{darkgreen}{0.089}& $-$ & $-$ & $-$ &\textcolor{darkgreen}{74.13} &\textcolor{darkgreen}{0.431}  &\textcolor{darkgreen}{0.618} \\ \cdashline{2-17}
& DSG \cite{DSG} & 20.15 & 0.207 &  0.885 & 26.35 &0.135& 0.862 &  31.89 &0.240&0.832& 27.60 & 0.217 & 0.920 &28.52 &0.201  &0.890\\ 
& CMI+DSG (ours) & \textcolor{darkgreen}{20.03} & \textcolor{darkgreen}{0.201} &  \textcolor{darkgreen}{0.890} & \textcolor{darkgreen}{26.02} &\textcolor{darkgreen}{0.127}& \textcolor{darkred}{0.860} &  \textcolor{darkgreen}{31.55} &\textcolor{darkgreen}{0.231}&\textcolor{darkgreen}{0.838}& \textcolor{darkgreen}{27.41} & \textcolor{darkgreen}{0.208} & \textcolor{darkgreen}{0.925} &\textcolor{darkgreen}{28.11} &\textcolor{darkgreen}{0.191}  &\textcolor{darkgreen}{0.892} \\
\midrule
\midrule
\multirow{7}{*}{ImageNet} & DPS \cite{DPS} & 35.87 &0.303& 0.739 & 38.82 &0.262& 0.794 & 62.72 &0.444& 0.706 &56.08 &0.389 &0.634& 50.66 & 0.337 & 0.781\\ 
&  CMI+DPS (ours) & \textcolor{darkgreen}{35.27} &\textcolor{darkgreen}{0.280}& \textcolor{darkgreen}{0.755} & \textcolor{darkgreen}{37.21} &\textcolor{darkgreen}{0.244}& \textcolor{darkgreen}{0.796} & \textcolor{darkgreen}{60.72} & \textcolor{darkgreen}{0.432} & \textcolor{darkgreen}{0.708} & \textcolor{darkgreen}{55.36} &\textcolor{darkgreen}{0.368} &\textcolor{darkgreen}{0.640}& \textcolor{darkgreen}{50.05} & \textcolor{darkgreen}{0.330} & \textcolor{darkred}{0.780}\\  \cdashline{2-17}
& $\Pi$GDM \cite{song2023pseudoinverse} & 41.82 &0.356& 0.705 & 42.26 &0.284& 0.752 & 59.79 &0.425& 0.717 &54.18 &0.373 &0.675& 54.26 & 0.352 & 0.765\\ 
& CMI+$\Pi$GDM (ours) & \textcolor{darkgreen}{38.41} &\textcolor{darkgreen}{0.334}& \textcolor{darkgreen}{0.721} & \textcolor{darkgreen}{40.90} &\textcolor{darkgreen}{0.277}& \textcolor{darkgreen}{0.772} & \textcolor{darkgreen}{59.55} & \textcolor{darkgreen}{0.410} & \textcolor{darkgreen}{0.719} &\textcolor{darkgreen}{53.85} &\textcolor{darkgreen}{0.366} & \textcolor{darkred}{0.673}& \textcolor{darkgreen}{52.22} & \textcolor{darkgreen}{0.350} & \textcolor{darkgreen}{0.782}\\ \cdashline{2-17}
& MCG \cite{MCG} & 39.19& 0.414&  0.546&  39.74 &0.330& 0.633  &95.04& 0.550& 0.441& $-$ & $-$ & $-$ &144.5 & 0.637 &0.227\\
& CMI+MCG (ours) & \textcolor{darkgreen}{37.18}& \textcolor{darkgreen}{0.379}&  \textcolor{darkgreen}{0.587}&  \textcolor{darkgreen}{38.24} &\textcolor{darkgreen}{0.310}& \textcolor{darkgreen}{0.685}  &\textcolor{darkgreen}{83.26}& \textcolor{darkgreen}{0.504}& \textcolor{darkgreen}{0.492}& $-$ & $-$ & $-$ &\textcolor{darkgreen}{124.2} & \textcolor{darkgreen}{0.606} &\textcolor{darkgreen}{0.341}\\ \cdashline{2-17}
& DSG \cite{DSG} & 32.41 & 0.218 &  0.753 & 33.27 &0.251& 0.811 &  56.44 &0.392&0.720& 52.01 & 0.348 & 0.643 &47.40 &0.318  &0.785\\ 
& CMI+DSG (ours) & \textcolor{darkgreen}{32.24} & \textcolor{darkgreen}{0.202} &  \textcolor{darkgreen}{0.757} & \textcolor{darkgreen}{33.11} &\textcolor{darkgreen}{0.239}& \textcolor{darkgreen}{0.813} &  \textcolor{darkgreen}{56.08} &\textcolor{darkgreen}{0.381}&\textcolor{darkgreen}{0.728}& \textcolor{darkgreen}{52.00} & \textcolor{darkgreen}{0.331} & \textcolor{darkgreen}{0.669} &\textcolor{darkgreen}{47.08} &\textcolor{darkgreen}{0.302}  &\textcolor{darkgreen}{0.801} \\
\bottomrule
\end{tabular}}
\vskip -0.2in
\end{table*}

\section{Experiments} \label{sec:exp}
Here, we demonstrate the impact of incorporating our CMI term into state-of-the-art inverse problem solvers with DMs, showing improvements both quantitatively and qualitatively.

\subsection{Quantitative Results.}\label{sec:Quantitative}
\noindent $\bullet$ \textbf{Experimental Setup.}  
Following \cite{DPS,dou2024diffusion}, we conduct experiments on the FFHQ 256$\times$256 \cite{karras2019style} and ImageNet 256$\times$256 datasets \cite{deng2009imagenet}, using 1k validation images scaled to $[0, 1]$. To ensure fairness, we adopt the settings from \cite{DPS}, with Gaussian noise (\(\sigma=0.05\)) added to the measurements. Inference uses \(N = 1000\) time steps, employing pre-trained score models from \cite{DPS} for FFHQ and \cite{dhariwal2021diffusion} for ImageNet.

The measurement models are based on those in \cite{DPS}:  
($i$) For box-type inpainting, a \(128 \times 128\) region is masked, while random-type inpainting masks 90\% of all pixels across RGB channels;  
($ii$) For super-resolution (SR), bicubic downsampling is applied;  
($iii$) For Gaussian blur, a \(61 \times 61\) kernel with a standard deviation of 5.0 is used;  
($iv$) For motion blur, randomly generated \(61 \times 61\) kernels (intensity 0.5) are convolved with ground truth images, using the code in \cite{motionblur}.

\noindent $\bullet$ \textbf{Benchmark Methods.}  
The benchmark methods used in our evaluation include DPS \cite{DPS}, $\Pi$GDM \cite{song2023pseudoinverse}, manifold constrained gradients (MCG) \cite{MCG}, and diffusion with
spherical Gaussian constraint (DSG) \cite{DSG}. For a fair comparison, we consistently apply the same pre-trained score function across all methods.

\noindent $\bullet$ \textbf{Evaluation Metrics.}  
We evaluate performance using metrics from \cite{DPS}:  
($i$) LPIPS \cite{zhang2018unreasonable},  
($ii$) FID \cite{heusel2017gans}, and  
($iii$) SSIM \cite{wang2004image}.  
All experiments run on a single A100 GPU.

\noindent $\bullet$ \textbf{Results.}
The results for both datasets are summarized in \cref{tab:quant}, showcasing that the integration of CMI with the benchmark methods substantially improves their performance in almost all the tasks.

\begin{figure}[!h]  
\centering\includegraphics[width=1\columnwidth]{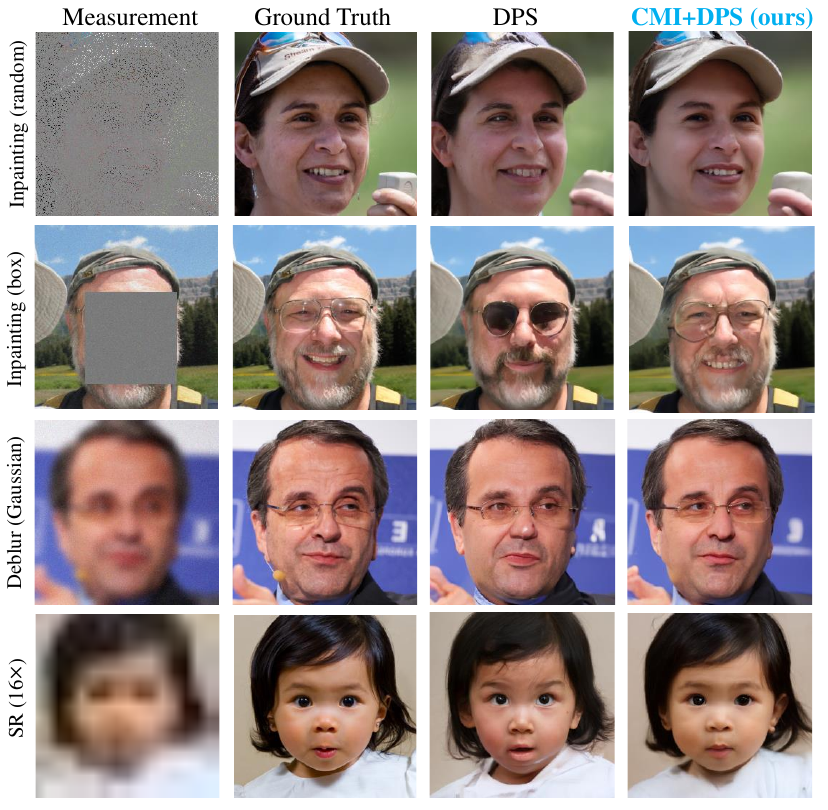} 
\vskip -0.2in
    \caption{Qualitative results on FFHQ dataset.} \label{fig:faces}
\vskip -0.2in  
\end{figure}

\subsection{Qualitative Results} \label{sec:Qualitative}
Lastly, we present a visual comparison of reconstructed images using DPS and CMI+DPS. Four images from the FFHQ test dataset are randomly selected and corrupted using the measurement methods in \cref{sec:Quantitative}, with a modification: applying \(16 \times\) super-resolution for a better observation. The reconstructed images are displayed in \cref{fig:faces}, with each row corresponding to a different measurement method. As observed, CMI+DPS produces reconstructions that more closely resemble the ground truth than DPS.

\section{Conclusion}
In this work, we have addressed the challenges of solving inverse problems using DMs by proposing a method that maximizes the conditional mutual information $\mathrm{I}(\boldsymbol{x}_0; \boldsymbol{y} | \boldsymbol{x}_t)$. This approach ensures that intermediate signals in the reverse path are generated to retain maximum information about the measurements, ultimately improving the quality of the reconstructed signal. Our method integrates seamlessly with existing state-of-the-art approaches, enhancing their performance across various inverse problems, including inpainting, deblurring, and super-resolution. The results demonstrated that this information-theoretic perspective provides both qualitative and quantitative improvements.

\bibliographystyle{IEEEtran}
\bibliography{egbib}

\end{document}